\documentclass[letterpaper]{article} 
\usepackage{aaai2026}  
\usepackage{times}  
\usepackage{helvet}  
\usepackage{courier}  
\usepackage[hyphens]{url}  
\usepackage{graphicx} 
\usepackage{natbib}  
\usepackage{caption} 
\usepackage{multirow}
\usepackage{amsmath,amsfonts}
\usepackage{booktabs}

\usepackage{newfloat}
\usepackage{listings}

\usepackage{algorithm}
\usepackage{algpseudocode}
\DeclareCaptionStyle{ruled}{labelfont=normalfont,labelsep=colon,strut=off} 
\floatstyle{ruled}
\newfloat{listing}{tb}{lst}{}
\floatname{listing}{Listing}
\title{Unveiling the Fragility of Vision-Language Models: Multi-Modal Adversarial Synergy via Texture-Constrained Perturbations and Cross-Modal Optimization}

\author{
    Xiang Fang\textsuperscript{\rm 1}, Wanlong Fang\textsuperscript{\rm 2}, Changshuo Wang\textsuperscript{\rm 3}\thanks{Corresponding Author. }
}
\affiliations{
    \textsuperscript{\rm 1}School of Software Engineering,
Huazhong University of Science and Technology\\ 
    \textsuperscript{\rm 2}Nanyang Technological University, Singapore\\ 
    \textsuperscript{\rm 3}University College London 

     xfang9508@gmail.com,  wanlongfang@gmail.com, wangchangshuo1@gmail.com
}

\begin{document}
\maketitle
\begin{abstract}
Large Vision-Language Models (LVLMs) have transformed multi-modal understanding, excelling in tasks like image captioning and visual question answering by integrating visual and textual inputs. However, their robustness against adversarial attacks—particularly those exploiting both modalities—remains underexplored, posing risks to critical applications like autonomous driving and content moderation. Existing attacks focus on single modalities or require impractical white-box access, limiting their real-world relevance. In this paper, we introduce \textit{Multi-Modal Adversarial Synergy}, a groundbreaking framework that crafts universal, black-box multi-modal attacks against LVLMs. MMAS simultaneously generates a texture scale-constrained Universal Adversarial Perturbation  for images and a learnable prompt perturbation for text, optimized jointly using only model queries. The image perturbation, bounded by an $\ell_{\infty}$-norm, leverages wavelet-based texture constraints to ensure imperceptibility and robustness across diverse visual inputs. The text perturbation, constrained by an $\ell_2$-norm in the embedding space, maintains semantic coherence while steering outputs toward a target. A novel cross-modal regularization term aligns the perturbations’ gradient directions, enhancing their synergistic impact and transferability across tasks and models. Extensive experiments show the strong universal adversarial capabilities of our proposed attack with prevalent LVLMs.

\end{abstract}    

\section{Introduction}
\label{sec:intro}

The rapid advancement of Large Vision-Language Models (LVLMs), such as CLIP \cite{radford2021learning,liang2025from,liang2023knowledge,tang2023m3net,tang2024divide,tang2022learning,tang2020blockmix,tang2025connecting}, has revolutionized multi-modal learning, enabling seamless integration of visual and textual data for tasks ranging from image captioning \cite{zeng2024meacap} to visual question answering (VQA) \cite{sima2024drivelm,liu2023exploring,wang2025taylor,fang2026towardsicml,kuai2026dynamic,wang2025point,fang2025your,zhang2025monoattack,fang2023hierarchical,liu2024towards,yang2025eood,fang2022multi,fang2026cogniVerse,lei2025exploring,fang2023you,wang2025dypolyseg,fang2025hierarchical,yan2026fit,fang2025adaptive,wang2026topadapter,cai2025imperceptible,fang2026slap,wang2026reasoning,fang2026immuno,wang2026biologically,fang2026disentangling,wang2025reducing,fang2026advancing,wang2026from,liu2023conditional,liu2026attacking,fang2026rethinking,wang2025seeing,fang2026towards,fang2025multi,fang2024fewer,liu2024pandora,fang2024multi,fang2025turing,fang2024not,liu2023hypotheses,fang2024rethinking,liu2024unsupervised,fang2023annotations,xiong2024rethinking,fang2021unbalanced,wang2025prototype,zhang2025manipulating,fang2026align,tang2024reparameterization,fang2025adaptivetai,tang2025simplification,fang2021animc,cai2026towards,fang2020v,fang2020double}. These models leverage vast pre-training datasets and sophisticated architectures to achieve remarkable generalization across diverse applications. However, their increasing deployment in real-world systems—such as autonomous vehicles \cite{cui2024drive,wang2025point,wang2026reasoning,wang2025dypolyseg,wang2026biologically,wang2025taylor}, content moderation \cite{liu2024content,ma2025efficient,ma2024safe,ma2024neural}, and medical diagnostics \cite{myrzashova2024safeguarding}—raises critical concerns about their robustness to adversarial attacks \cite{xing2025towards}. Adversarial perturbations \cite{li2022learning}, subtle alterations to inputs designed to mislead machine learning models, have been extensively studied in unimodal contexts, such as image classification \cite{rao2021global} and natural language processing \cite{min2023recent}. Yet, the multi-modal nature of LVLMs introduces a new frontier: how resilient are these models to coordinated attacks across both vision and language modalities?

Recent works have begun to explore adversarial vulnerabilities in LVLMs, revealing alarming weaknesses \cite{dai2025data}. For instance, \cite{luo2024image} demonstrated that carefully crafted text prompts can mislead LVLMs across multiple tasks, exploiting the models' sensitivity to prompt variations. Similarly, \cite{liu2024pandora} introduced universal adversarial patches that disrupt image understanding in a task-agnostic manner, while \cite{yin2023vlattack} extended this to black-box settings, showing transferability across models. In the vision domain, \cite{huang2024texture} proposed texture scale-constrained perturbations, highlighting how structured noise can enhance attack robustness. Despite these advances, existing approaches predominantly focus on single-modality attacks, leaving a critical gap in understanding how joint image-text perturbations might amplify adversarial effects. This gap is particularly pressing given the interdependent processing of vision and language in LVLMs, where cross-modal interactions could be exploited to craft more potent and practical attacks.

The implications of this vulnerability are profound. In safety-critical applications, such as autonomous driving \cite{zhao2024autonomous}, an attacker could pair a perturbed road sign image with a misleading textual instruction to cause catastrophic misinterpretation. In content moderation, subtle image-text manipulations could bypass filters, allowing harmful content to proliferate. Moreover, the universal nature of LVLMs—designed to handle arbitrary image-text pairs—suggests that effective attacks must generalize across tasks and inputs, a challenge unmet by task-specific or modality-isolated methods. Existing multi-modal attack methods \cite{abdullakutty2021review} remain limited to white-box settings with access to model gradients, rendering them impractical for real-world scenarios where only query access is available. Thus, there is an urgent need for a practical, black-box, multi-modal attack framework that exploits the synergy between image and text perturbations while ensuring universality and transferability.

In this paper, we introduce \textit{Multi-Modal Adversarial Synergy (MMAS)}, a novel framework to craft universal adversarial attacks against LVLMs. MMAS simultaneously generates a texture scale-constrained Universal Adversarial Perturbation (UAP) for images and a learnable prompt perturbation for text, optimized jointly in a black-box setting using only model queries. Our approach builds upon insights from prior works: we adapt the texture-constrained UAP concept  to ensure visual imperceptibility and robustness, draw on the prompt perturbation strategy  for text attacks, and incorporate the universal and black-box optimization principles. However, MMAS transcends these foundations by introducing a key innovation: a cross-modal regularization term that aligns the image and text perturbations, enhancing their combined efficacy and transferability across diverse tasks and inputs.

Our method operates under realistic constraints. The image perturbation, bounded by an $\ell_{\infty}$-norm, leverages wavelet-based texture scales to maintain perceptual similarity while disrupting visual features universally. The text perturbation, optimized in the embedding space with an $\ell_2$-norm constraint, ensures semantic coherence while steering the LVLM toward a target output. By jointly optimizing these perturbations with a query-based gradient approximation, MMAS achieves a practical attack that requires no internal model knowledge—a significant departure from white-box methods. Furthermore, the cross-modal regularization encourages synergy between modalities, addressing the limitation of prior works where image and text attacks were designed independently, often resulting in suboptimal performance when combined.

We evaluate MMAS on a range of LVLMs, including CLIP and Flamingo, across tasks such as image captioning, VQA, and text-guided image classification. Our results demonstrate that MMAS achieves higher attack success rates and better transferability compared to single-modality baselines and naive multi-modal combinations. For example, in a targeted attack scenario, MMAS can manipulate an LVLM to consistently misclassify a ``stop sign'' image paired with a ``proceed'' prompt as ``go'', with perturbations imperceptible to human observers. These findings underscore the vulnerability of LVLMs to coordinated multi-modal attacks and highlight the need for robust defenses.

Our contributions can be summarized as follows: 1) We propose MMAS, the first black-box, universal multi-modal attack framework for LVLMs, integrating texture scale-constrained image perturbations and learnable text prompt perturbations. 2) We introduce a novel cross-modal regularization term to enhance the synergy between image and text perturbations, improving attack efficacy and transferability. 3) We provide extensive evaluations showing MMAS's superiority over existing methods, offering new insights into the multi-modal vulnerabilities of LVLMs.

\section{Related Work}
\label{sec:related}


 Adversarial attacks in the vision domain originated with seminal works  \cite{goodfellow2014explaining,akhtar2018threat,zhang2020adversarial}, which introduced gradient-based perturbations to mislead image classifiers. Subsequent studies refined these attacks for practicality and generality. For instance, \cite{moosavi2017universal} pioneered Universal Adversarial Perturbations (UAPs), single noise patterns effective across multiple images, enhancing attack scalability. More recently, \cite{huang2024texture} proposed texture scale-constrained perturbations, leveraging wavelet transforms to craft robust, visually coherent noise that withstands image transformations. While effective against vision-only models, these methods do not address the multi-modal nature of LVLMs, where text inputs play a critical role. 


The intersection of vision and language in LVLMs has spurred initial multi-modal attack explorations \cite{wallace2019universal}. \cite{lapid2023see} combined image noise with text alterations in a white-box setting, achieving targeted mispredictions in CLIP-like models. However, its dependence on model internals restricts real-world applicability. \cite{yin2023vlattack} advanced this by developing task-agnostic perturbations in a black-box framework, using query-based optimization to approximate gradients. While promising, it focuses on image perturbations with static text, missing the opportunity to jointly optimize both modalities. \cite{luo2024image} and \cite{liu2024pandora} hint at multi-modal potential by pairing their respective text and image attacks, but these combinations are ad hoc, lacking a unified optimization strategy. No prior work has systematically addressed the synergy between image and text perturbations in a universal, black-box context, leaving LVLMs’ full vulnerabilities underexplored.

\begin{figure*}[t!]
\centering
\includegraphics[width=\textwidth]{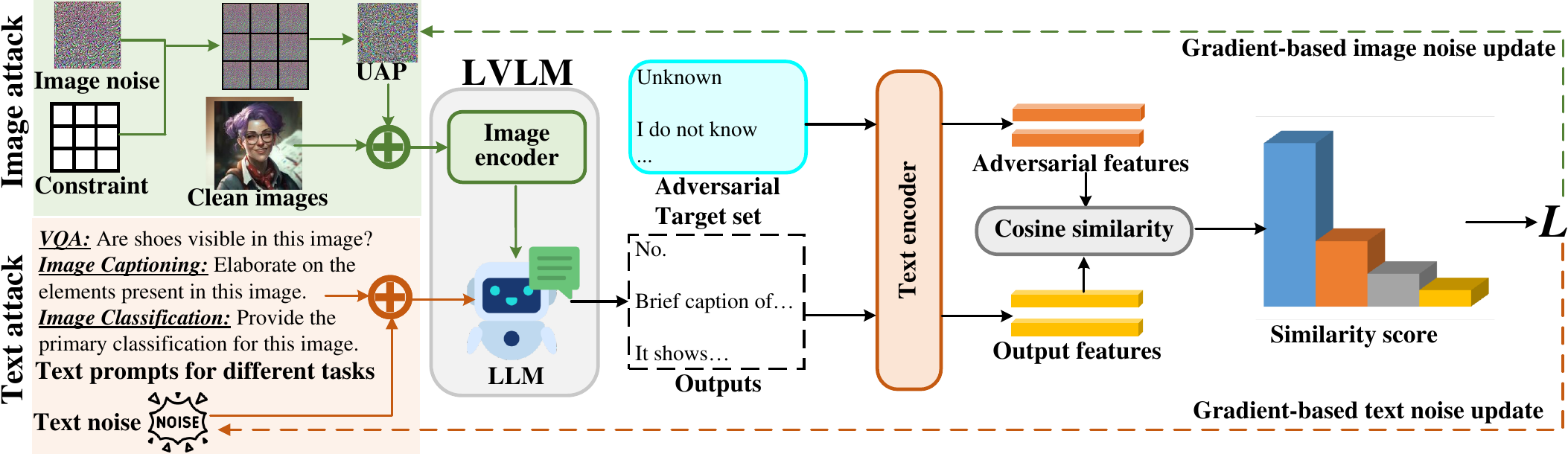}
\caption{Overview of our proposed method. The framework initializes a texture scale-constrained Universal Adversarial Perturbation (UAP) for images and a learnable prompt perturbation for text, followed by joint optimization using query-based Projected Gradient Descent with cross-modal regularization to align perturbations. The resulting universal attack is evaluated on unseen image-text pairs across diverse tasks and LVLMs, achieving high efficacy and transferability.
Best viewed in color.
}
\label{fig:pipeline}
\end{figure*}

\section{Methodology}

In this section, we present \textit{Multi-Modal Adversarial Synergy (MMAS)}, a novel framework to craft multi-modal adversarial attacks against Large Vision-Language Models (LVLMs). Our approach simultaneously generates a texture scale-constrained Universal Adversarial Perturbation (UAP) for images and a learnable prompt perturbation for text, aiming to mislead LVLMs across diverse tasks and prompts. The method operates in a practical black-box setting, relying solely on model queries, and introduces a cross-modal regularization term to enhance attack transferability. Figure~\ref{fig:pipeline} illustrates the MMAS pipeline, comprising initialization, joint optimization, and evaluation stages.


\noindent \textbf{Problem formulation.} Consider an LVLM $f_{\theta}(\mathbf{v}, \mathbf{t})$ that processes a clean image $\mathbf{v} \in \mathbb{R}^{H \times W \times C}$ and a textual prompt $\mathbf{t} \in \mathcal{T}$ to produce an output $\mathbf{y}$. Our goal is to craft an adversarial image $\mathbf{v}' = \mathbf{v} + \delta_v$ and an adversarial prompt $\mathbf{t}' = \mathbf{t} + \delta_t$, where $\delta_v$ is the image perturbation and $\delta_t$ is the text perturbation, such that the LVLM consistently outputs a predefined target text $\mathbf{y}'$ (targeted attack) across various tasks, i.e., $f_{\theta}(\mathbf{v}', \mathbf{t}') \mapsto \mathbf{y}'$. We impose constraints $\|\delta_v\|_{\infty} \leq \epsilon_v$ and $\|\delta_t\|_2 \leq \epsilon_t$ to ensure imperceptibility and semantic coherence, respectively. The attack is universal, meaning $\delta_v$ and $\delta_t$ are task-agnostic and applicable to any input pair $(\mathbf{v}, \mathbf{t})$.

\noindent \textbf{Threat model.}
In this paper, we explore the scenario of attacking real-world LVLM models. In this paper, the assumption is that the attacker has no knowledge of the victim model, including its parameters, training procedure, original training data, etc.
In particular, unlike in white-/gray-box attacks, we cannot access the model's gradient information to train perturbations through back-propagation, nor can we, as in black-box attacks, obtain confidence scores/logits from the model outputs.
\subsection{Texture Scale-Constrained UAP for Image Attack}
\label{subsec:image_attack}
To craft a Universal Adversarial Perturbation (UAP) featuring category-specific local textures, one straightforward approach is to divide the UAP into numerous small sections, each tailored with a distinct texture. Given that UAP creation relies on gradient back-propagation across its entire scale, adjusting the gradient for each segment in a detailed manner becomes intricate and challenging. To address this, we suggest reversing the perspective and streamlining the task: starting with a category-specific local texture patch, how can we assemble a larger patch—matching the dimensions of the training images—as the UAP? A practical solution is to tile the patch into a broader expanse, leveraging a standard image processing technique that incurs minimal effort.
We design $\delta_v$ as a Universal Adversarial Perturbation (UAP) constrained by texture scales to enhance its robustness and transferability across images and tasks. Unlike traditional UAPs, our perturbation leverages multi-scale texture features extracted via a wavelet transform to maintain visual coherence while maximizing adversarial impact.

Let $\mathcal{W}(\mathbf{v})$ denote the wavelet decomposition of the clean image $\mathbf{v}$, yielding coefficients at multiple scales $\{s_1, s_2, \ldots, s_L\}$. We constrain $\delta_v$ to align with the texture patterns at scale $s_k$ by projecting it onto the wavelet subspace: $\delta_v = \sum_{k=1}^L s_k \cdot \mathcal{W}^{-1}(\mathbf{W}_k \odot \mathcal{W}(\mathbf{v})),$
where $\mathcal{W}^{-1}$ is the inverse wavelet transform; $\mathbf{W}_k$ is a binary mask selecting coefficients at scale $k$; $\odot$ denotes the Hadamard product; $\alpha_k$ are learnable weights. To ensure universality, $\delta_v$ is initialized randomly within the bound $\|\delta_v\|_{\infty} \leq \epsilon_v$ and optimized across a diverse set of images $\mathcal{V} = \{\mathbf{v}_1, \ldots, \mathbf{v}_N\}$. The texture constraint prevents overfitting to specific image content, enhancing cross-image and cross-task transferability.

\subsection{Learnable Prompt Perturbation for Text Attack}
\label{subsec:text_attack}
In many vision-language applications, the  text input is quite short, which makes the existing LVLMs vulnerable to attack. For example, the average length of the text in  VQAv2 \cite{goyal2017making} and RefCOCO  is 6.21 and 3.57, respectively. Moreover, some words are nonsense, making it unnecessary to design a new approach for attacking the text modality.
Therefore, we introduce a learnable prompt perturbation $\delta_t$ to the text input $\mathbf{t}$. Unlike fixed prompts, $\delta_t$ is optimized in the embedding space of the LVLM's text encoder $g_{\phi}$. For a prompt $\mathbf{t}$, its embedding is $e_{\mathbf{t}} = g_{\phi}(\mathbf{t})$. The adversarial prompt embedding becomes: $e_{\mathbf{t}'} = e_{\mathbf{t}} + \delta_t,$
where $\delta_t$ is constrained by $\|\delta_t\|_2 \leq \epsilon_t$. During optimization, $\delta_t$ is updated to maximize the language modeling loss away from the original output (non-targeted) or minimize it toward the target $\mathbf{y}'$ (targeted). This perturbation is applied only during training, and the final attack uses the optimized $\delta_t$ universally across prompts $\mathcal{T} = \{\mathbf{t}_1, \ldots, \mathbf{t}_M\}$.

\subsection{Joint Optimization by Cross-Modal Regularization}
\label{subsec:optimization}
In numerous instances, altering just images or texts proves challenging for success, since modifying a single modality often fails to sever the link between visuals and captions. To tackle this issue, we introduce a brand-new  joint optimization strategy combining image and text perturbations based on a novel cross-modal regularization module. The objective balances the attack success with modality synergy. For a targeted attack, we define the language modeling loss as 
\small
\begin{equation}
    \mathcal{L}(\mathbf{y}, \mathbf{y}') = -\log P(\mathbf{y}' | f_{\theta}(\mathbf{v}', \mathbf{t}')).
\end{equation}\normalsize
The optimization problem is:
\small
\begin{equation}
    \min_{\delta_v, \delta_t} \frac{1}{|\mathcal{V}| |\mathcal{T}|} \sum_{\mathbf{v} \in \mathcal{V}} \sum_{\mathbf{t} \in \mathcal{T}} \left[ \mathcal{L}(f_{\theta}(\mathbf{v} + \delta_v, \mathbf{t} + \delta_t), \mathbf{y}') + \lambda \mathcal{R}(\delta_v, \delta_t) \right],\nonumber
\end{equation}\normalsize
where $\|\delta_v\|_{\infty} \leq \epsilon_v$ and $\|\delta_t\|_2 \leq \epsilon_t$, where $\lambda$ is a hyperparameter, $\mathcal{R}(\delta_v, \delta_t)$ means the cross-modal regularization: $\mathcal{R}(\delta_v, \delta_t) = \left\| \nabla_{\delta_v} \mathcal{L} \cdot \nabla_{\delta_t} \mathcal{L} \right\|_2$.
This term encourages alignment between the gradient directions of $\delta_v$ and $\delta_t$, ensuring that image and text perturbations reinforce each other, improving attack efficacy across prompts and tasks.
Since we operate in a black-box setting, we approximate gradients using a query-based method similar to \cite{yin2023vlattack}. For $\delta_v$, we sample noise $\eta_v \sim \mathcal{U}(-\epsilon_v, \epsilon_v)$ and estimate: 
\small
\begin{align}
    \nabla_{\delta_v} \mathcal{L} &\approx [\frac{\mathcal{L}(f_{\theta}(\mathbf{v} + \delta_v + \eta_v, \mathbf{t}'), \mathbf{y}') }{\|\eta_v\|_2} \nonumber\\
    &-\frac{ \mathcal{L}(f_{\theta}(\mathbf{v} + \delta_v, \mathbf{t}'), \mathbf{y}')}{\|\eta_v\|_2}]\cdot \frac{\eta_v}{\|\eta_v\|_2}.
\end{align}\normalsize
Similarly, for $\delta_t$, we use noise $\eta_t \sim \mathcal{N}(0, \epsilon_t^2)$ in the embedding space. The perturbations are updated via Projected Gradient Descent (PGD): $\delta_v^{t+1} = \text{Proj}_{\epsilon_v} \left( \delta_v^t - \alpha_v \cdot \text{sign}(\nabla_{\delta_v} \mathcal{L}) \right)$ and $\delta_t^{t+1} = \text{Proj}_{\epsilon_t} \left( \delta_t^t - \alpha_t \cdot \frac{\nabla_{\delta_t} \mathcal{L}}{\|\nabla_{\delta_t} \mathcal{L}\|_2} \right)$,
where $\alpha_v$ and $\alpha_t$ are step sizes, and $\text{Proj}$ ensures constraints are met. The texture scale constraint on $\delta_v$ is reapplied after each update.


The MMAS algorithm is detailed in Algorithm~\ref{alg:mmas}. It initializes $\delta_v$ and $\delta_t$, iteratively optimizes them using the joint objective, and evaluates the universal attack on unseen inputs. This framework ensures a novel, robust, and universal multi-modal attack, leveraging cross-modal synergy for enhanced transferability across LVLMs.

\begin{algorithm}[t]
\caption{Multi-Modal Adversarial Synergy (MMAS)}
\label{alg:mmas}
\small
\begin{algorithmic}[1]
\Require LVLM $f_{\theta}$, target text $\mathbf{y}'$, image set $\mathcal{V}$, prompt set $\mathcal{T}$, bounds $\epsilon_v$, $\epsilon_t$, step sizes $\alpha_v$, $\alpha_t$, iterations $T$, regularization weight $\lambda$
\Ensure Adversarial perturbations $\delta_v$, $\delta_t$
\State Initialize $\delta_v \sim \mathcal{U}(-\epsilon_v, \epsilon_v)$ with texture scale constraint, $\delta_t \sim \mathcal{N}(0, \epsilon_t^2)$
\For{$t = 1$ to $T$}
    \State Sample $(\mathbf{v}, \mathbf{t})$ from $\mathcal{V} \times \mathcal{T}$
    \State Compute $\mathcal{L} = \mathcal{L}(f_{\theta}(\mathbf{v} + \delta_v, \mathbf{t} + \delta_t), \mathbf{y}')$
    \State Estimate gradients $\nabla_{\delta_v} \mathcal{L}$ and $\nabla_{\delta_t} \mathcal{L}$ via queries
    \State Compute $\mathcal{R}(\delta_v, \delta_t) = \left\| \nabla_{\delta_v} \mathcal{L} \cdot \nabla_{\delta_t} \mathcal{L} \right\|_2$
    \State Update $\delta_v^{t+1}$ and $\delta_t^{t+1}$ using PGD with $\mathcal{L} + \lambda \mathcal{R}$
    \State Apply texture scale constraint to $\delta_v^{t+1}$
\EndFor
\State \Return  $\delta_v$, $\delta_t$
\end{algorithmic}
\end{algorithm}

\section{Experiments}
\label{sec:experiments}


\begin{table}[t!]
\centering
\scriptsize
\setlength\tabcolsep{0.05cm}
\begin{tabular}{cc|cccc|c}
\toprule
Target Model & Method & Classification & Captioning & {VQA\textsubscript{general}} & {VQA\textsubscript{specific}} & Overall \\ \midrule
\multicolumn{7}{c}{Dataset: MS-COCO} \\ \midrule
\multirow{4}{*}{LLaVA} & Clean & 0.316 & 0.423 & 0.327 & 0.358 & 0.356 \\
                        & TA-UAP & 0.834 & 0.845 & 0.807 & 0.845 & 0.833 \\
                        & TC-UAP & 0.804 & 0.815 & 0.777 & 0.815 & 0.803 \\
                        & Ours(Full) & \textbf{0.884} & \textbf{0.895} & \textbf{0.857} & \textbf{0.895} & \textbf{0.883} \\ \cmidrule(lr){1-7}
\multirow{4}{*}{MiniGPT-4} & Clean & 0.386 & 0.401 & 0.421 & 0.440 & 0.412 \\
                           & TA-UAP & 0.826 & 0.842 & 0.834 & 0.873 & 0.844 \\
                           & TC-UAP & 0.796 & 0.812 & 0.804 & 0.843 & 0.814 \\
                           & Ours(Full) & \textbf{0.876} & \textbf{0.892} & \textbf{0.884} & \textbf{0.923} & \textbf{0.894} \\ \cmidrule(lr){1-7}
\multirow{4}{*}{Flamingo} & Clean & 0.412 & 0.427 & 0.449 & 0.483 & 0.443 \\
                          & TA-UAP & 0.847 & 0.805 & 0.831 & 0.843 & 0.832 \\
                          & TC-UAP & 0.810 & 0.788 & 0.819 & 0.830 & 0.812 \\
                          & Ours(Full) & \textbf{0.895} & \textbf{0.877} & \textbf{0.892} & \textbf{0.916} & \textbf{0.895} \\ \cmidrule(lr){1-7}
\multirow{4}{*}{BLIP-2} & Clean & 0.413 & 0.422 & 0.487 & 0.526 & 0.462 \\
                        & TA-UAP & 0.799 & 0.746 & 0.804 & 0.847 & 0.799 \\
                        & TC-UAP & 0.769 & 0.716 & 0.774 & 0.817 & 0.769 \\
                        & Ours(Full) & \textbf{0.849} & \textbf{0.796} & \textbf{0.854} & \textbf{0.897} & \textbf{0.849} \\ \midrule
\multicolumn{7}{c}{Dataset: DALLE-3} \\ \midrule
\multirow{4}{*}{LLaVA} & Clean & 0.310 & 0.420 & 0.472 & 0.495 & 0.424 \\
                        & TA-UAP & 0.785 & 0.846 & 0.804 & 0.853 & 0.822 \\
                        & TC-UAP & 0.755 & 0.816 & 0.774 & 0.823 & 0.792 \\
                        & Ours(Full) & \textbf{0.835} & \textbf{0.896} & \textbf{0.854} & \textbf{0.903} & \textbf{0.872} \\ \cmidrule(lr){1-7}
\multirow{4}{*}{MiniGPT-4} & Clean & 0.302 & 0.325 & 0.367 & 0.392 & 0.347 \\
                           & TA-UAP & 0.843 & 0.836 & 0.825 & 0.848 & 0.838 \\
                           & TC-UAP & 0.813 & 0.806 & 0.795 & 0.818 & 0.808 \\
                           & Ours(Full) & \textbf{0.893} & \textbf{0.886} & \textbf{0.875} & \textbf{0.898} & \textbf{0.888} \\ \cmidrule(lr){1-7}
\multirow{4}{*}{Flamingo} & Clean & 0.406 & 0.452 & 0.427 & 0.479 & 0.441 \\
                          & TA-UAP & 0.822 & 0.849 & 0.852 & 0.887 & 0.852 \\
                          & TC-UAP & 0.792 & 0.819 & 0.822 & 0.857 & 0.822 \\
                          & Ours(Full) & \textbf{0.872} & \textbf{0.899} & \textbf{0.902} & \textbf{0.937} & \textbf{0.903} \\ \cmidrule(lr){1-7}
\multirow{4}{*}{BLIP-2} & Clean & 0.325 & 0.378 & 0.410 & 0.439 & 0.388 \\
                        & TA-UAP & 0.821 & 0.799 & 0.875 & 0.886 & 0.845 \\
                        & TC-UAP & 0.788 & 0.754 & 0.838 & 0.847 & 0.807 \\
                        & Ours(Full) & \textbf{0.866} & \textbf{0.873} & \textbf{0.910} & \textbf{0.938} & \textbf{0.897} \\ \midrule
\multicolumn{7}{c}{Dataset: VQAv2} \\ \midrule
\multirow{4}{*}{LLaVA} & Clean & 0.401 & 0.423 & 0.475 & 0.487 & 0.447 \\
                        & TA-UAP & 0.793 & 0.762 & 0.809 & 0.847 & 0.803 \\
                        & TC-UAP & 0.763 & 0.732 & 0.779 & 0.817 & 0.773 \\
                        & Ours(Full) & \textbf{0.843} & \textbf{0.812} & \textbf{0.859} & \textbf{0.897} & \textbf{0.853} \\ \cmidrule(lr){1-7}
\multirow{4}{*}{MiniGPT-4} & Clean & 0.402 & 0.414 & 0.519 & 0.543 & 0.470 \\
                           & TA-UAP & 0.846 & 0.835 & 0.848 & 0.884 & 0.853 \\
                           & TC-UAP & 0.816 & 0.805 & 0.818 & 0.854 & 0.823 \\
                           & Ours(Full) & \textbf{0.896} & \textbf{0.885} & \textbf{0.898} & \textbf{0.934} & \textbf{0.903} \\ \cmidrule(lr){1-7}
\multirow{4}{*}{Flamingo} & Clean & 0.349 & 0.382 & 0.396 & 0.25 & 0.344 \\
                          & TA-UAP & 0.829 & 0.786 & 0.827 & 0.865 & 0.827 \\
                          & TC-UAP & 0.799 & 0.768 & 0.803 & 0.841 & 0.803 \\
                          & Ours(Full) & \textbf{0.869} & \textbf{0.875} & \textbf{0.852} & \textbf{0.899} & \textbf{0.874} \\ \cmidrule(lr){1-7}
\multirow{4}{*}{BLIP-2} & Clean & 0.312 & 0.340 & 0.357 & 0.398 & 0.352 \\
                        & TA-UAP & 0.776 & 0.825 & 0.846 & 0.874 & 0.830 \\
                        & TC-UAP & 0.746 & 0.795 & 0.816 & 0.844 & 0.800 \\
                        & Ours(Full) & \textbf{0.826} & \textbf{0.875} & \textbf{0.896} & \textbf{0.924} & \textbf{0.880} \\ \bottomrule
\end{tabular}
\caption{Attack performance on various LVLM models. }
\label{tab:Main_table}
\end{table}

\begin{figure}[t!]
    \centering
    \includegraphics[width=\columnwidth]{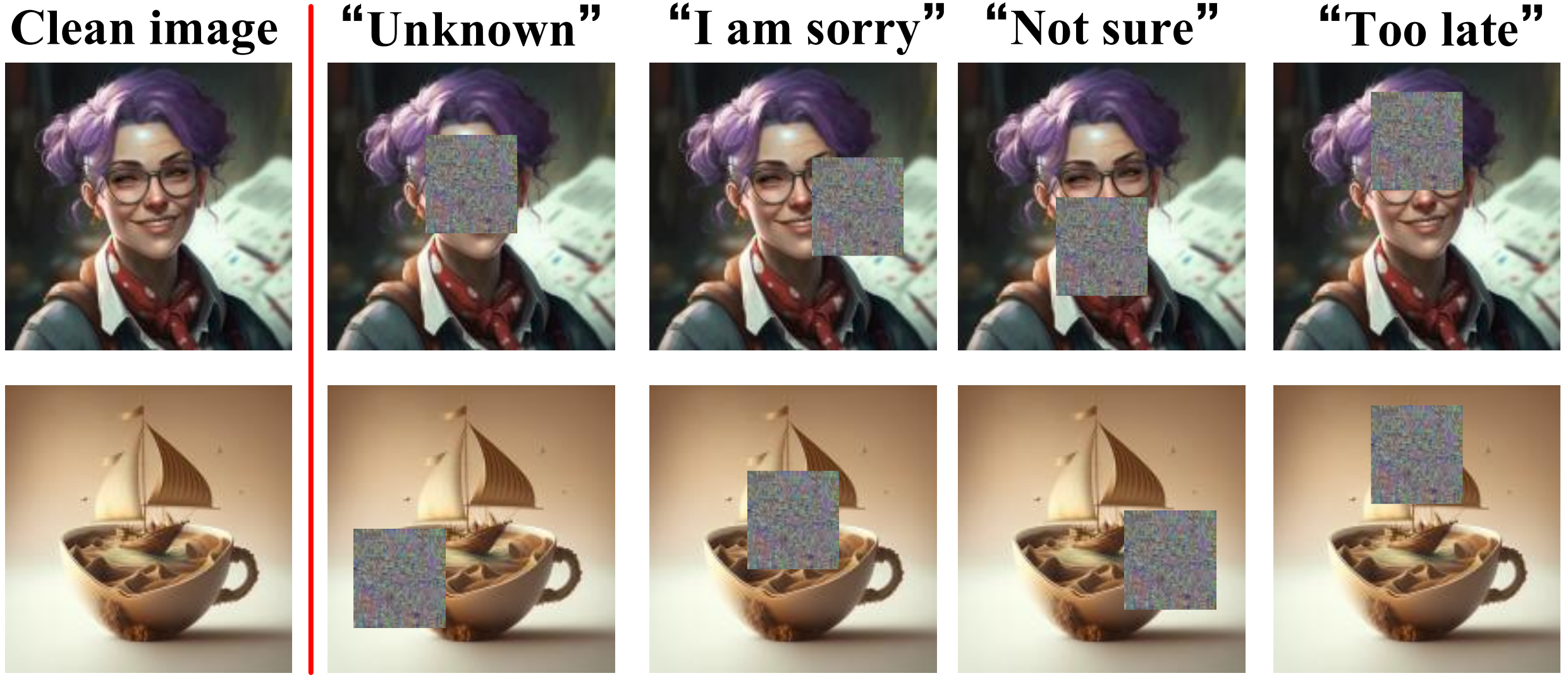}
    \caption{Visualization  on the  universal adversarial attack.}
    \label{fig:visualization}
\end{figure}

\begin{table}[t!]
\scriptsize
\setlength\tabcolsep{0.25cm}
\centering
\begin{tabular}{c|c|ccc|c}
\toprule
Method & Attack  & LLaVA & BLIP-2 & MiniGPT-4 & Mean \\ \midrule
MF-Attack  & black-box & 0.590 & 0.681 & 0.668 & 0.646 \\
Ours & universal  & \textbf{0.879} & \textbf{0.894}  & \textbf{0.745} & \textbf{0.839} \\ \bottomrule
\end{tabular}
\caption{Comparison with  MF-Attack \cite{zhao2024evaluating}. For a fair comparison, experiments are conducted on the same ImageNet-1k dataset \cite{deng2009imagenet} in the VQA task.}
\label{tab:compare_to_NIPS}
\end{table}

\begin{table}[t!]
\centering
\scriptsize
\setlength\tabcolsep{0.08cm}
\begin{tabular}{c|c|cccc|c}
\toprule
Method & Attack  & Classification & Captioning & {VQA\textsubscript{general}} & {VQA\textsubscript{specific}} & Overall \\ \midrule
CroPA  & white-box  & 0.75 & 0.72 & {0.90}&0.96 & 0.83 \\
Ours & universal  & \textbf{0.86}& 0.78& 0.95& 0.99& 0.89\\ \bottomrule
\end{tabular}
\caption{Comparison with  CroPA \cite{luo2024image}. For  fair comparison, we follow CroPA to evaluate the same ASR metric on   OpenFlamingo  \cite{awadalla2023openflamingo} and MS-COCO.}
\label{tab:compare_to_CroPA}
\end{table}

\begin{table}[t!]
    \centering
    \scriptsize
\setlength\tabcolsep{0.25cm}
    \begin{tabular}{l|cc|cc|cccccc}
        \hline
        \multirow{2}{*}{ Attacker} & \multicolumn{2}{c|}{Gemini-2.0} & \multicolumn{2}{c|}{GPT-4o} & \multicolumn{2}{c}{Claude-3.5}\\
        & SS$\uparrow$ & ASR$\uparrow$ & SS$\uparrow$ & ASR$\uparrow$ & SS$\uparrow$ & ASR$\uparrow$  \\
        \hline
        PGD  &0.084& 0.013& 0.076& 0.006& 0.098& 0.024  \\
        CroPA  & 0.132& 0.020& 0.084& 0.011& 0.114& 0.085  \\
        \textbf{Ours} & \textbf{0.576}& \textbf{0.413}& \textbf{0.508}& \textbf{0.379}& \textbf{0.645}& \textbf{0.496} \\
        \hline
    \end{tabular}
        \caption{\small Attack performance on the generated adversarial examples from LLaVA to commercial VLMs, where ``SS'' means ``semantic similarity'' and ``ASR'' means ``Attack Success Rate''.}
    \label{shangye}
\end{table}

\begin{table}[t!]
 \scriptsize
\setlength\tabcolsep{0.25cm}
    \centering
    \begin{tabular}{ccl}
        \toprule
         Image &  Method & \multicolumn{1}{c}{ LVLM-Output} \\
        \midrule
        \multirow{1}{*}{\includegraphics[width=0.13\textwidth,height=1.8cm]{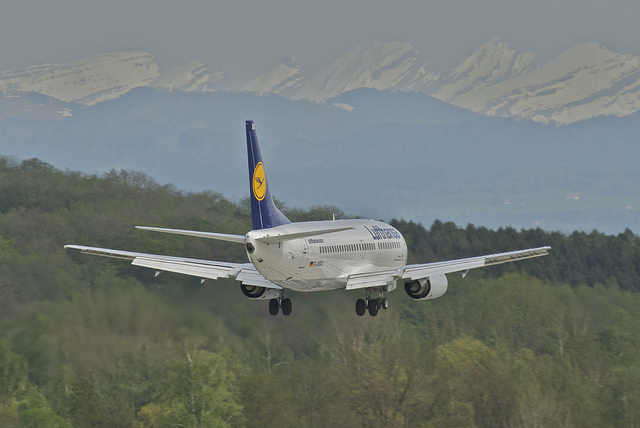}} & No Attack&  A plane flying between mountains  \\
        \cmidrule(lr){2-3}
        & PGD &  An airplane and a bird  flying  \\
        \cmidrule(lr){2-3}
        & CroPA & Two people repairing the plane \\
        \cmidrule(lr){2-3}
        & Ours&  A tiger eating meat \\
        \midrule
        \multirow{1}{*}{\includegraphics[width=0.13\textwidth,height=2cm]{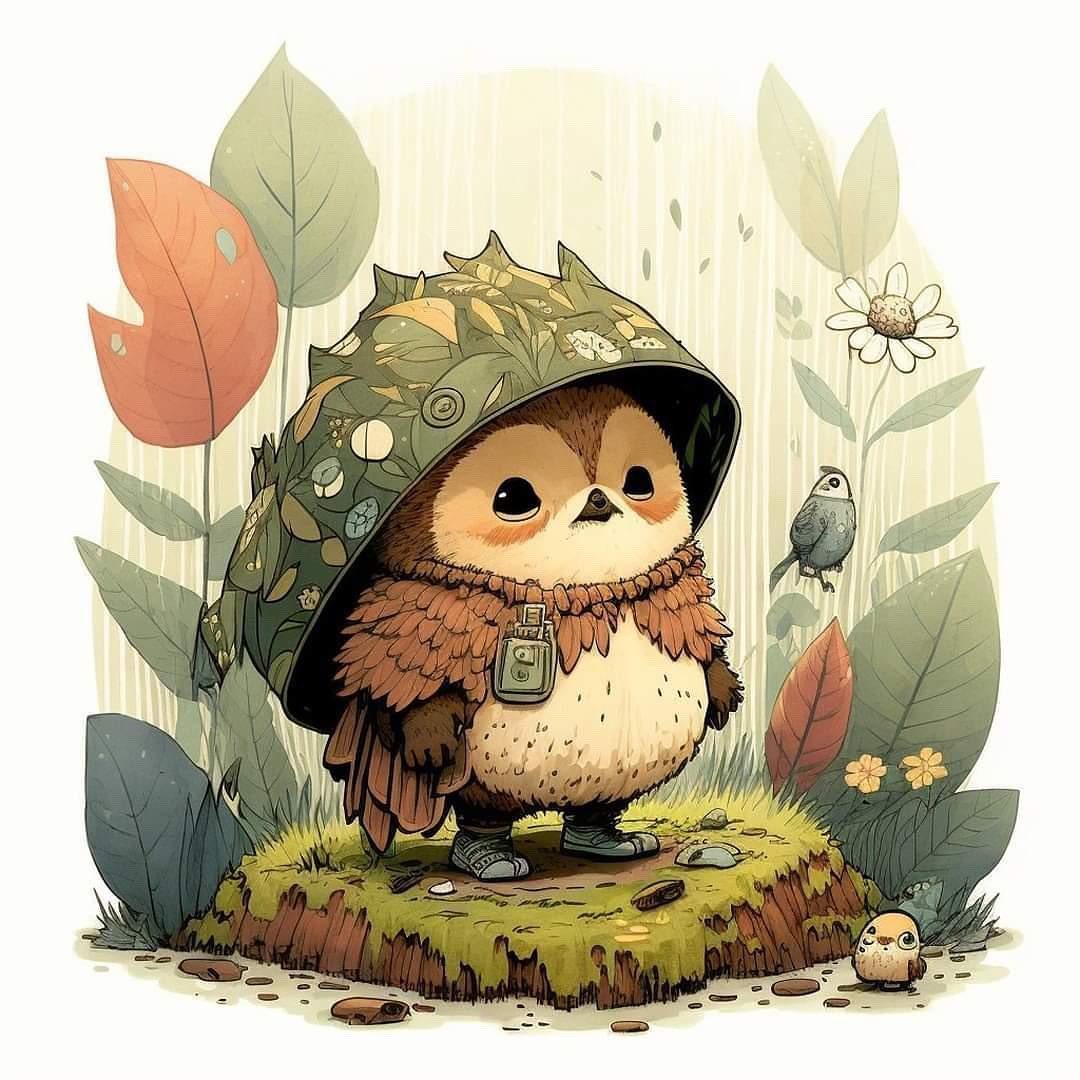}} & No Attack &  An owl standing on a hill with birds  \\
        \cmidrule(lr){2-3}
        & PGD &  An owl eating leaves \\
        \cmidrule(lr){2-3}
        & CroPA& A little bird standing on a branch \\
        \cmidrule(lr){2-3}
        & Ours &  Two hedgehogs fighting \\
        \bottomrule
        
    \end{tabular}
            \caption{Cases of attack results (top: MS-COCO, bottom: DALLE-3) against LLaVA in different methods.} \label{tab:llavacase}
\end{table}
\begin{table}[t!]
\centering
 \scriptsize
\setlength\tabcolsep{0.05cm}
\begin{tabular}{cccccccccc}
\toprule
From & \multicolumn{1}{c|}{Transfer to} & Classification & Captioning & {VQA\textsubscript{general}} & {VQA\textsubscript{specific}} & Overall \\ \midrule
\multicolumn{7}{c}{Transferability across Different Datasets (on Flamingo)} \\ \midrule
\multirow{3}{*}{MS-COCO} & \multicolumn{1}{c|}{MS-COCO} & 0.895 & 0.877 & 0.892 & 0.916 & 0.895 \\
~ & \multicolumn{1}{c|}{DALLE-3} & 0.850 & 0.832 & 0.847 & 0.871 & 0.850 \\
 & \multicolumn{1}{c|}{VQAv2} & 0.845 & 0.827 & 0.842 & 0.866 & 0.845 \\ \cmidrule{1-7}
\multirow{3}{*}{DALLE-3} & \multicolumn{1}{c|}{MS-COCO} & 0.835 & 0.862 & 0.865 & 0.900 & 0.866 \\
~ & \multicolumn{1}{c|}{DALLE-3} & 0.872 & 0.899 & 0.902 & 0.937 & 0.903 \\
 & \multicolumn{1}{c|}{VQAv2} & 0.835 & 0.862 & 0.865 & 0.900 & 0.866 \\ \cmidrule{1-7}
\multirow{3}{*}{VQAv2} & \multicolumn{1}{c|}{MS-COCO} & 0.832 & 0.838 & 0.815 & 0.862 & 0.837 \\
 & \multicolumn{1}{c|}{DALLE-3} & 0.832 & 0.838 & 0.815 & 0.862 & 0.837 \\
 ~ & \multicolumn{1}{c|}{VQAv2} & 0.869 & 0.875 & 0.852 & 0.899 & 0.874 \\
 \midrule
\multicolumn{7}{c}{Transferability across Different LVLM Models (on MS-COCO)} \\ \midrule
\multirow{4}{*}{LLaVA} & \multicolumn{1}{c|}{LLaVA} & 0.884 & 0.895 & 0.857 & 0.895 & 0.883 \\
~ & \multicolumn{1}{c|}{MiniGPT-4} & 0.854 & 0.865 & 0.827 & 0.865 & 0.853 \\
 & \multicolumn{1}{c|}{Flamingo} & 0.864 & 0.847 & 0.862 & 0.886 & 0.865 \\
 & \multicolumn{1}{c|}{BLIP-2} & 0.824 & 0.766 & 0.824 & 0.867 & 0.820 \\ \cmidrule{1-7}
\multirow{4}{*}{MiniGPT-4} & \multicolumn{1}{c|}{LLaVA} & 0.846 & 0.862 & 0.854 & 0.893 & 0.864 \\
~ & \multicolumn{1}{c|}{MiniGPT-4} & 0.876 & 0.892 & 0.884 & 0.923 & 0.894 \\
 & \multicolumn{1}{c|}{Flamingo} & 0.865 & 0.847 & 0.862 & 0.896 & 0.868 \\
 & \multicolumn{1}{c|}{BLIP-2} & 0.846 & 0.862 & 0.854 & 0.893 & 0.864 \\ \cmidrule{1-7}
\multirow{4}{*}{Flamingo} & \multicolumn{1}{c|}{LLaVA} & 0.865 & 0.847 & 0.862 & 0.886 & 0.865 \\
 & \multicolumn{1}{c|}{MiniGPT-4} & 0.855 & 0.862 & 0.854 & 0.893 & 0.866 \\
 ~ & \multicolumn{1}{c|}{Flamingo} & 0.895 & 0.877 & 0.892 & 0.916 & 0.895 \\
 & \multicolumn{1}{c|}{BLIP-2} & 0.825 & 0.766 & 0.824 & 0.867 & 0.821 \\ \cmidrule{1-7}
\multirow{4}{*}{BLIP-2} & \multicolumn{1}{c|}{LLaVA} & 0.819 & 0.766 & 0.824 & 0.867 & 0.819 \\
 & \multicolumn{1}{c|}{MiniGPT-4} & 0.826 & 0.762 & 0.824 & 0.867 & 0.820 \\
 & \multicolumn{1}{c|}{Flamingo} & 0.829 & 0.766 & 0.824 & 0.867 & 0.822 \\
 ~ & \multicolumn{1}{c|}{BLIP-2} & 0.849 & 0.796 & 0.854 & 0.897 & 0.849 \\
 \bottomrule
\end{tabular}
\caption{Investigation on the transferability across different datasets and LVLMs with semantic similarity score for evaluation.}
\label{tab:transferability}
\end{table}

\begin{figure}[t!]
    \centering
       \includegraphics[width=0.23\textwidth]{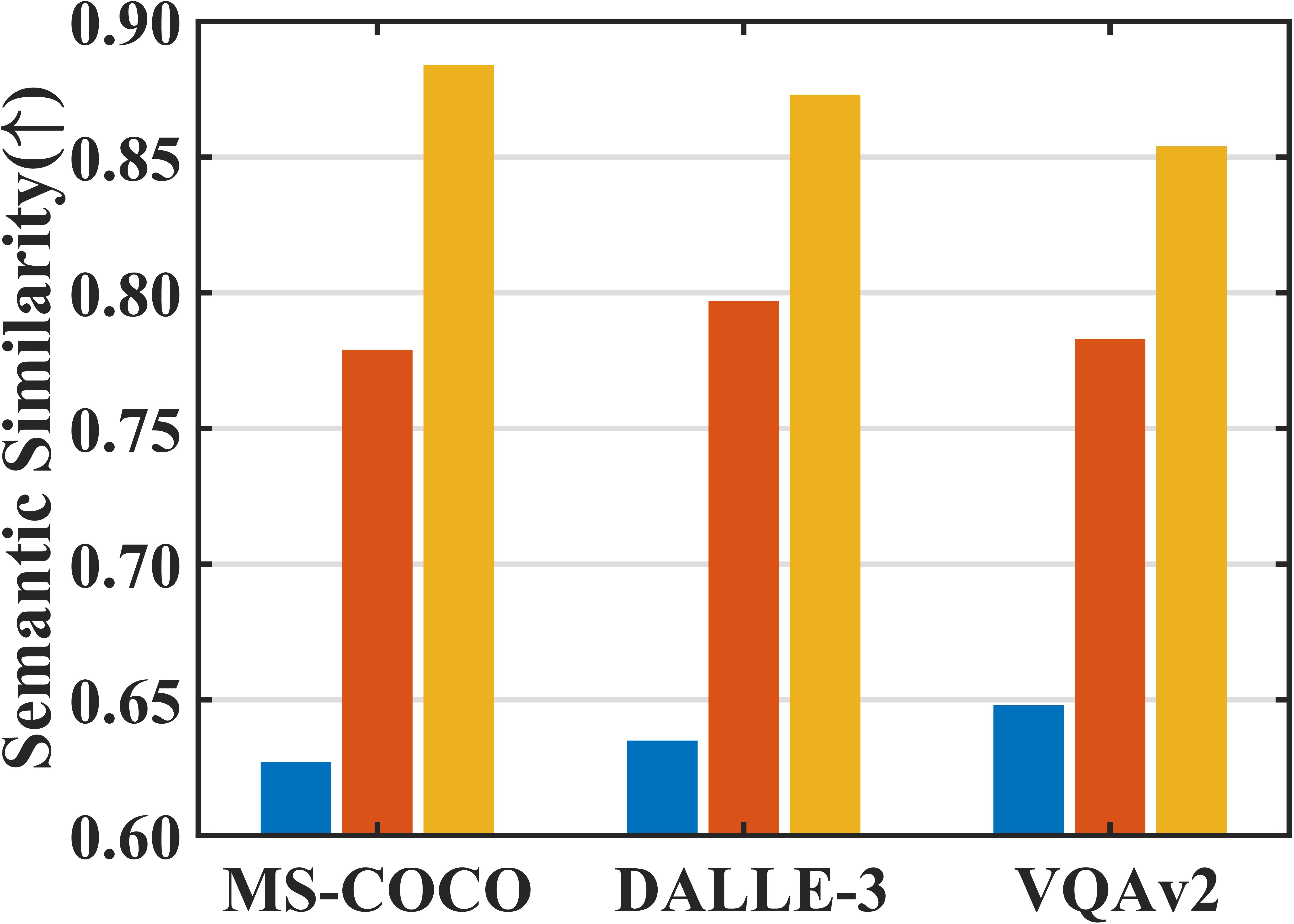}
\hspace{-0.05in}
\includegraphics[width=0.23\textwidth]{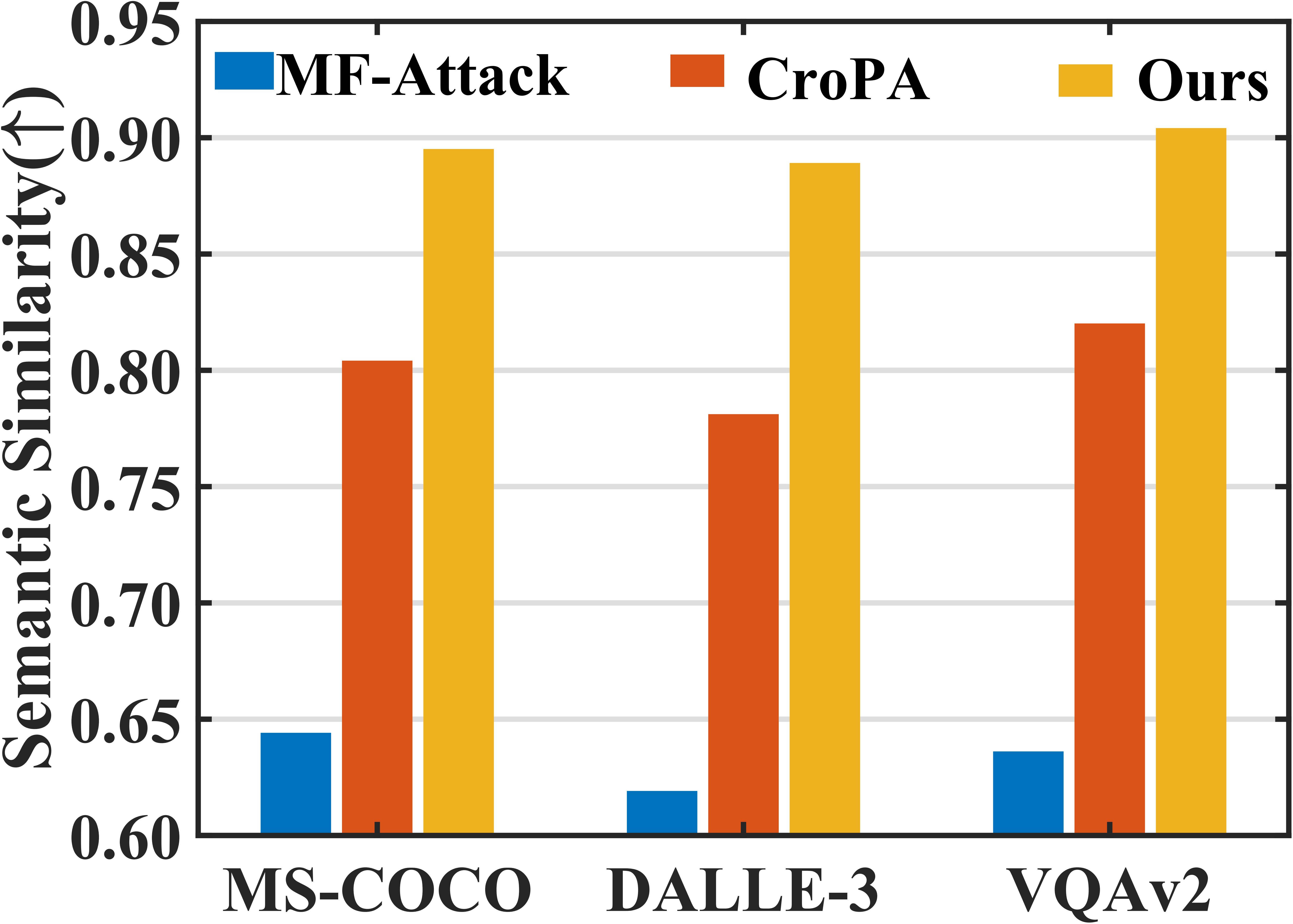}
\includegraphics[width=0.23\textwidth]{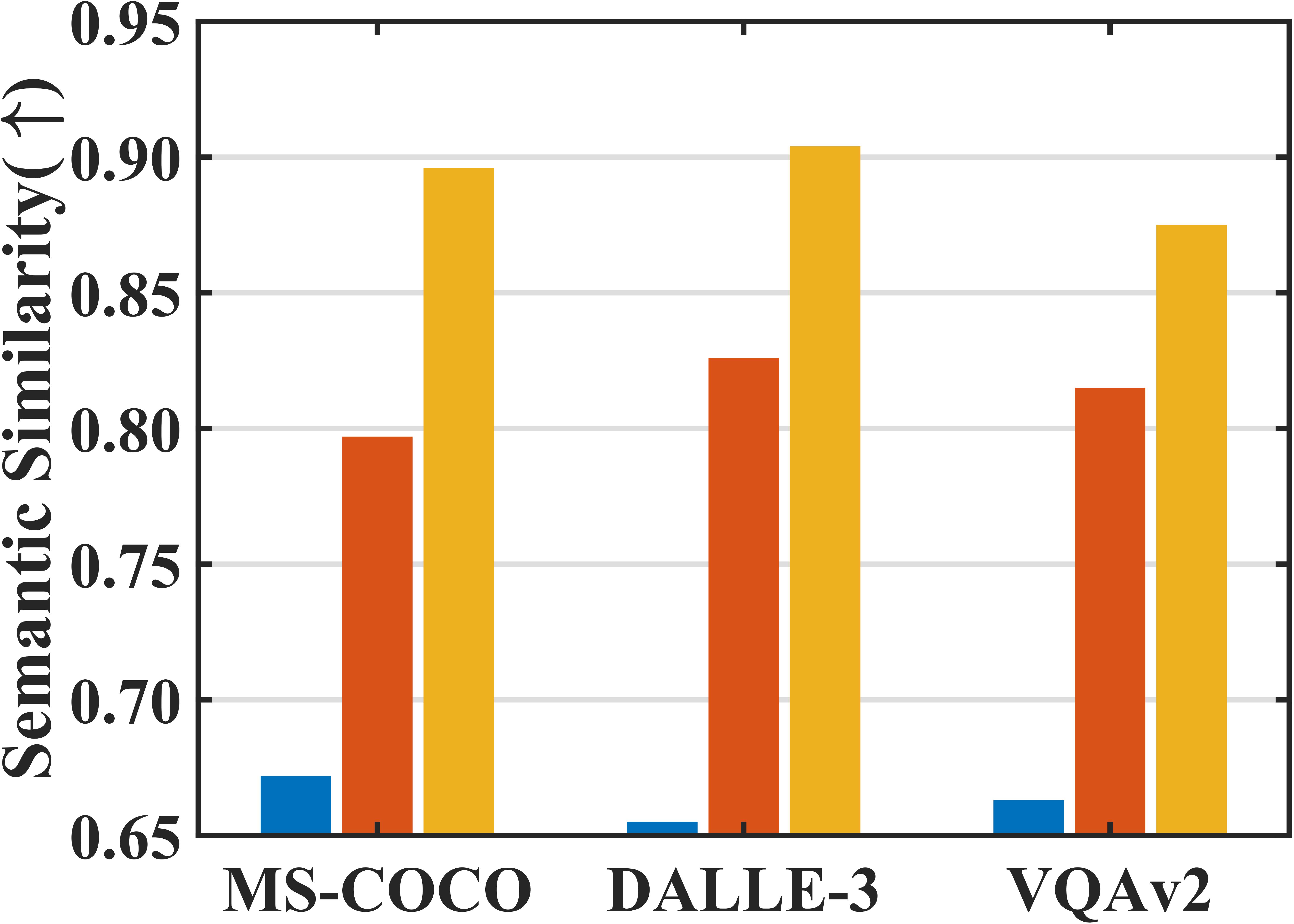}
\hspace{-0.05in}
\includegraphics[width=0.23\textwidth]{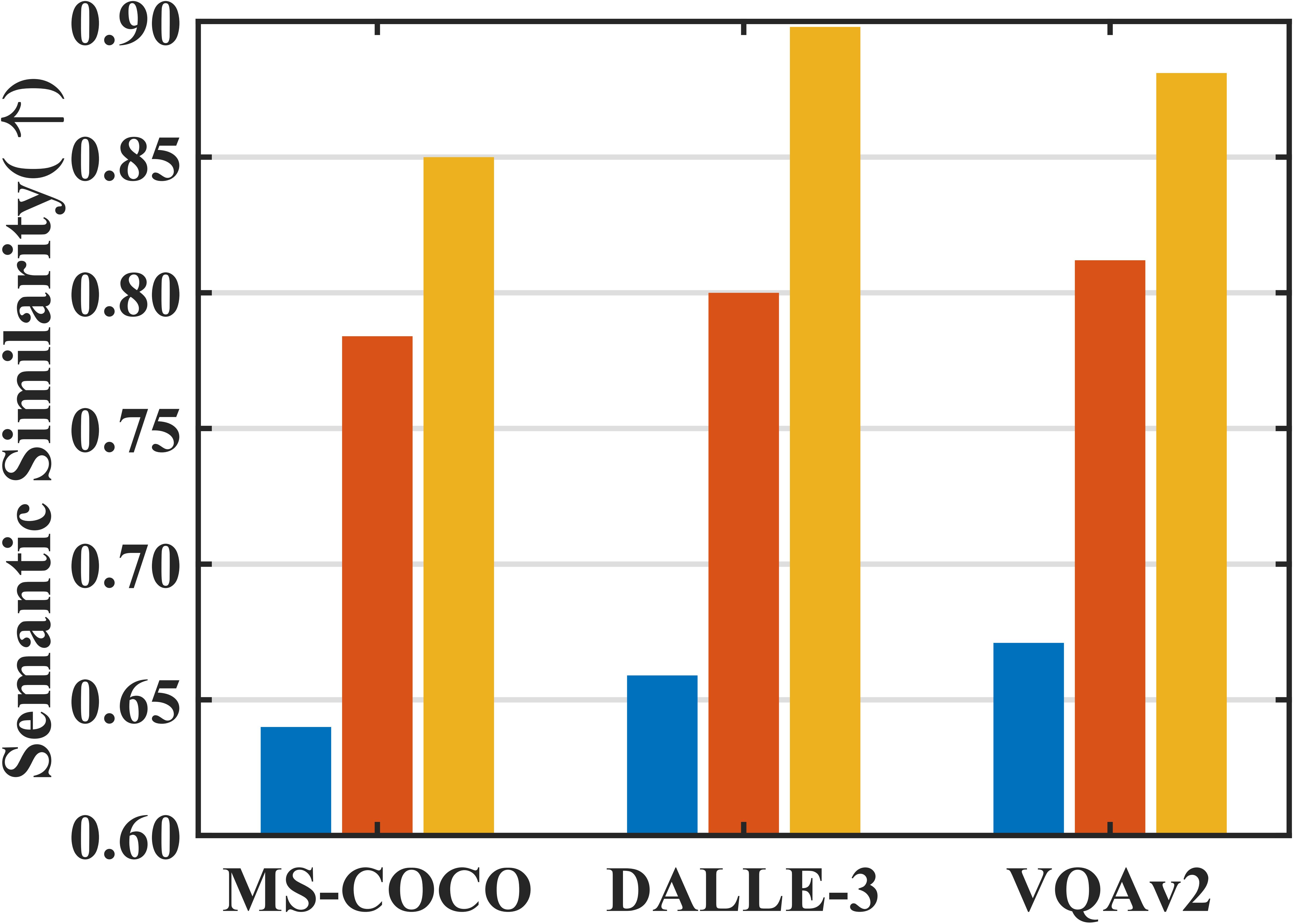}
    \caption{Performance comparison (Overall) with different LVLM attacks on different LVLM models across different datasets (Top left: LLaVA, Top right: MiniGPT-4, Bottom left: Flamingo, Bottom right: BLIP-2).}
    \label{fig:different_target}
\end{figure}
\begin{figure}[t!]
    \centering
       \includegraphics[width=0.23\textwidth]{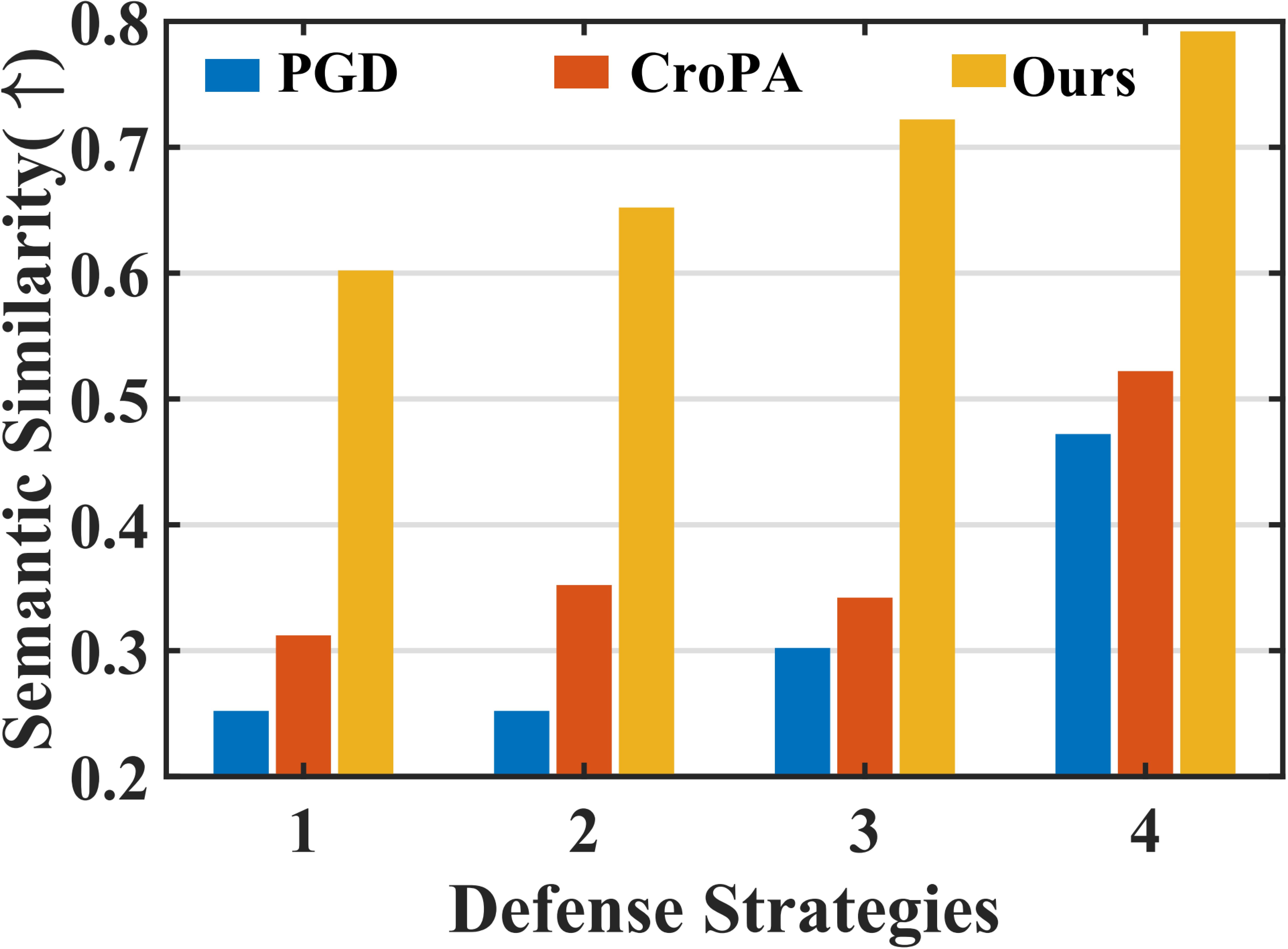}
\hspace{-0.05in}
\includegraphics[width=0.23\textwidth]{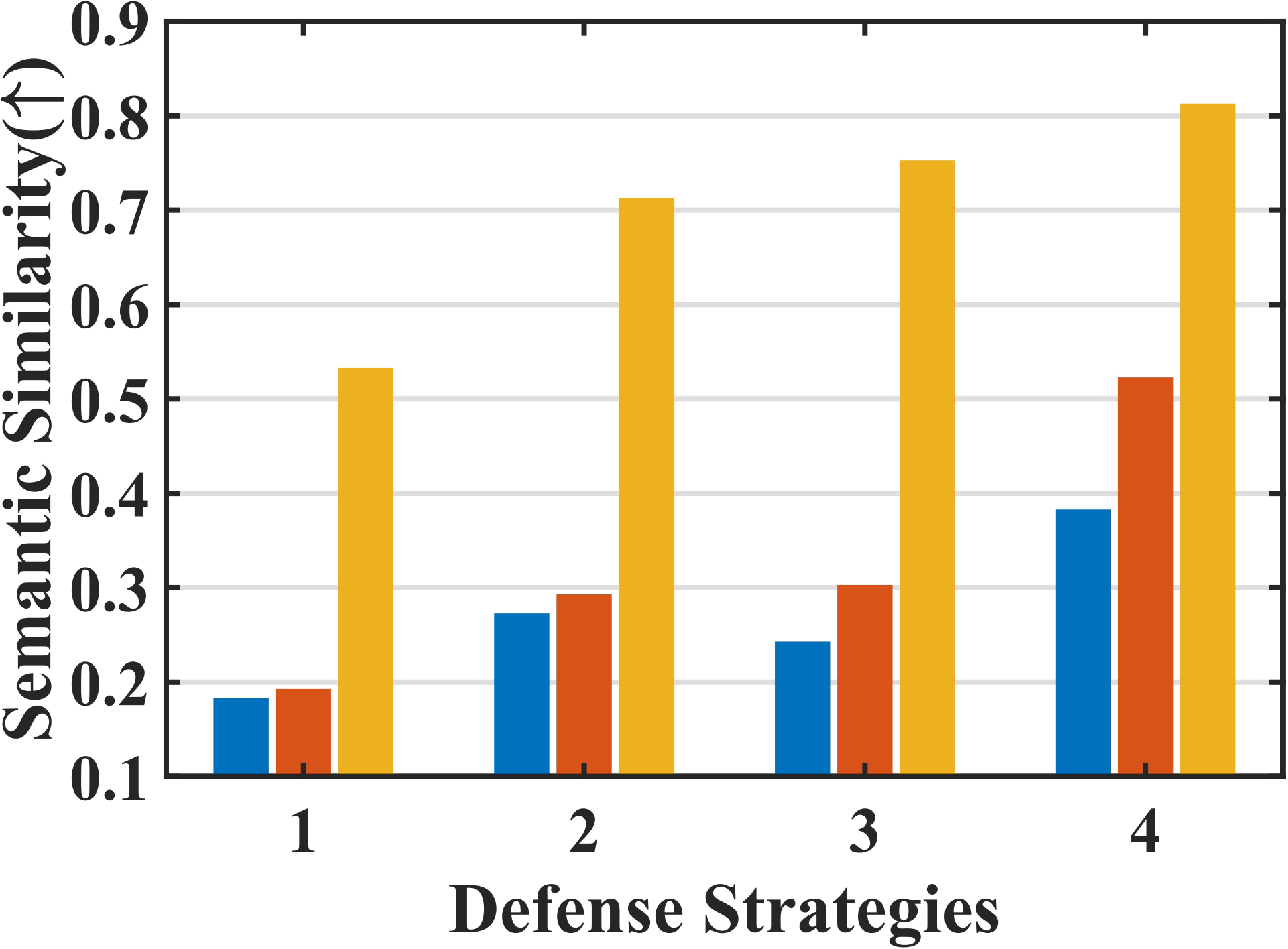}
    \caption{Investigation on the adversarial robustness against various defense methods on   MS-COCO (left) and DALLE-3 (right), tested on LLaVa model. Strategy 1: Diffusion Restoration \cite{nie2022diffusion};  strategy 2: Randomization \cite{xie2017mitigating};  strategy 3: JPEG Compression \cite{guo2017countering}; strategy 4: No Defense.}
    \label{fig:defense}
\end{figure}

\begin{table}[t!]
\centering
 \scriptsize
\setlength\tabcolsep{0.7cm}
\begin{tabular}{lccc}
\toprule
 Attacker & GPU Time ($\downarrow$) & GPU Memory ($\downarrow$) \\
\midrule
PGD  & 8 min & 14.3 GB \\
CroPA  & 17 min & 31.6 GB \\\hline
Ours & \textbf{6 min} & \textbf{13.5 GB} \\
\bottomrule
\end{tabular}
\caption{Complexity comparison on adversarial sample generation.} \label{complex}
\end{table}

\begin{table}[t!]
 \scriptsize
\setlength\tabcolsep{0.05cm}
\centering
\begin{tabular}{c|cccccccccccccccccccc}
\hline
{Model}&   Classification & Captioning & {VQA\textsubscript{general}} & {VQA\textsubscript{specific}} & Overall  \\\hline
w/o Text attack& 0.756& 0.752& 0.438& 0.829&0.693\\
w/o Image attack& 0.624& 0.741& 0.543& 0.816& 0.681\\
w/o Joint optimization& 0.796& 0.823&0.710& 0.841& 0.793\\\hline
Ours& \textbf{0.835} & \textbf{0.896} & \textbf{0.854} & \textbf{0.903} & \textbf{0.872} \\\hline
\end{tabular}
\caption{Ablation study on different  attacks, tested on the LLaVA model and DALLE-3 dataset.} \label{tab:ablation_attack}
\end{table}

\begin{table}[t!]
\centering
 \scriptsize
\setlength\tabcolsep{0.04cm}
\begin{tabular}{clccccc}
\hline
{Target Prompt} & {Method} & Classification & Captioning & {VQA\textsubscript{general}} & {VQA\textsubscript{specific}} & Overall \\
\hline
\multirow{4}{*}{{Unknown}} &
 Clean & 0.398 & 0.415 & 0.437 & 0.469 & 0.430 \\
 & TA-UAP & 0.837 & 0.795 & 0.821 & 0.833 & 0.822 \\
 & TC-UAP & 0.800 & 0.778 & 0.809 & 0.820 & 0.802 \\
 & Ours(Full) & \textbf{0.885} & \textbf{0.867} & \textbf{0.882} & \textbf{0.906} & \textbf{0.885} \\
\hline
\multirow{4}{*}{{I am sorry}} &
 Clean & 0.412 & 0.427 & 0.449 & 0.483 & 0.443 \\
 & TA-UAP & 0.847 & 0.805 & 0.831 & 0.843 & 0.832 \\
 & TC-UAP & 0.810 & 0.788 & 0.819 & 0.830 & 0.812 \\
 & Ours(Full) & \textbf{0.895} & \textbf{0.877} & \textbf{0.892} & \textbf{0.916} & \textbf{0.895} \\
\hline
\multirow{4}{*}{{Not sure}} &
 Clean & 0.405 & 0.420 & 0.442 & 0.475 & 0.436 \\
 & TA-UAP & 0.842 & 0.800 & 0.826 & 0.838 & 0.827 \\
 & TC-UAP & 0.805 & 0.783 & 0.814 & 0.825 & 0.807 \\
 & Ours(Full) & \textbf{0.890} & \textbf{0.872} & \textbf{0.887} & \textbf{0.911} & \textbf{0.890} \\
\hline
\multirow{4}{*}{{Very good}} &
 Clean & 0.425 & 0.438 & 0.415 & 0.452 & 0.433 \\
 & TA-UAP & 0.827 & 0.785 & 0.811 & 0.823 & 0.812 \\
 & TC-UAP & 0.790 & 0.768 & 0.799 & 0.810 & 0.792 \\
 & Ours(Full) & \textbf{0.875} & \textbf{0.857} & \textbf{0.872} & \textbf{0.896} & \textbf{0.875} \\
\hline
\multirow{4}{*}{{Too late}} &
 Clean & 0.385 & 0.410 & 0.428 & 0.460 & 0.421 \\
 & TA-UAP & 0.832 & 0.790 & 0.816 & 0.828 & 0.817 \\
 & TC-UAP & 0.795 & 0.773 & 0.804 & 0.815 & 0.797 \\
 & Ours(Full) & \textbf{0.880} & \textbf{0.862} & \textbf{0.877} & \textbf{0.901} & \textbf{0.880} \\
\hline
\multirow{4}{*}{{Metaphor}} &
 Clean & 0.375 & 0.395 & 0.408 & 0.435 & 0.403 \\
 & TA-UAP & 0.822 & 0.780 & 0.801 & 0.813 & 0.804 \\
 & TC-UAP & 0.785 & 0.763 & 0.789 & 0.800 & 0.784 \\
 & Ours(Full) & \textbf{0.870} & \textbf{0.852} & \textbf{0.862} & \textbf{0.886} & \textbf{0.868} \\
\hline
\end{tabular}
\caption{Targeted semantic similarity scores tested on Flamingo with different target texts on  MS-COCO.}
\label{tab:different_target}
\end{table}

\begin{table}[t!]
\centering
 \scriptsize
\setlength\tabcolsep{0.21cm}
\begin{tabular}{lccccccc}
\toprule
$s_k$ & Classification & Captioning & {VQA\textsubscript{general}} & {VQA\textsubscript{specific}} & Overall \\
\midrule
1 & 0.811 & 0.790 &0.835 & 0.857 &0.823  \\
2 & 0.825 & 0.822 &0.870& 0.884 & 0.850 \\
4 & \textbf{0.835} & \textbf{0.896} & \textbf{0.854} & \textbf{0.903} & \textbf{0.872}  \\
8 & 0.817 & 0.816 &0.859 & 0.870 &  0.841\\
\bottomrule
\end{tabular}
\caption{Ablation on scaling factor $s_k$ (LLaVA-1.5, DALLE-3).}
\label{tab:ablation_scale}
\end{table}

 \subsection{Experimental Setup}

\textbf{Models:} Following existing LVLM attack methods \cite{shayegani2023jailbreak,bailey2023image,dong2023robust,wang2023instructta,wang2024stop,luo2024image,tao2024imgtrojan,zhao2024evaluating}, we assess the current popular open-source LVLMs, including LLaVA \cite{liu2024visual}, MiniGPT-4 \cite{zhu2023minigpt}, Flamingo \cite{alayrac2022flamingo}, and BLIP-2 \cite{li2023blip}.
 To ensure a fair comparison of attack effectiveness, we utilize the same pretrained weights of LVLMs as used in previous studies \cite{liu2024visual,zhu2023minigpt,alayrac2022flamingo,li2023blip}.


\noindent \textbf{Datasets:} To accurately evaluate the attack methodologies, we utilized three challenging datasets to capture a range of visual and contextual diversity: MS-COCO \cite{lin2014microsoft}, VQAv2 \cite{goyal2017making}, and DALLE-3 \cite{ramesh2022hierarchical}. We also follow the existing works to construct these three datasets.
Specifically, we employ images from the test sets of the MS-COCO and VQAv2 to construct two multimodal datasets.
We also use captions from the MS-COCO validation set as prompts to generate corresponding images with DALLE-3 to form another dataset.
For the text input data, we follow the prompts used in previous work \cite{luo2024image} to build our text dataset, with detailed data presented in the supplementary material.


\noindent \textbf{Tasks:} We evaluated MMAS across three core LVLM tasks: 1) \textit{Image Classification}, where models assign a single-label descriptor (e.g., ``dog''); 2) \textit{Image Captioning}, generating a descriptive sentence (e.g., ``A dog sits on grass''); and 3) \textit{Visual Question Answering (VQA)}, answering yes/no or open-ended questions (e.g., ``Is there a dog?''). 

\noindent \textbf{Attack settings.} We craft targeted attacks, aiming to force LVLMs to output a predefined target (e.g., ``go'' for a stop sign image with a ``proceed'' prompt). Image perturbations are constrained by $\|\delta_v\|_{\infty} \leq 8/255$, and text perturbations by $\|\delta_t\|_2 \leq 0.5$ in the embedding space, ensuring imperceptibility. MMAS is optimized over 100 iterations with step sizes $\alpha_v = 0.01$, $\alpha_t = 0.005$, and regularization weight $\lambda = 0.1$, using a batch size of 16.

\noindent \textbf{Implementation details:} We implemented MMAS in PyTorch, using wavelet transforms from PyWavelets \cite{lee2019pywavelets} for texture constraints. Queries approximate gradients with 10 samples per iteration, simulating a black-box setting. MMAS was implemented with a base texture patch of size $64 \times 64$, tiled with scaling factor $s_k \in \{1, 2, 4, 8\}$. The perturbation magnitude was constrained to $\eta = 16/255$ (pixel range [0, 1]), ensuring near-imperceptibility. Optimization ran for $T = 70,000$ queries, with noise variance $\sigma = 0.01$, threshold $\theta = 0.55$, and weight scaling $\gamma = 5$. We used Sentence-BERT~\cite{reimers2019sentence} as the text encoder to compute cosine similarity between outputs and the target $\mathbf{r}^* = \text{``I am sorry''}$. Experiments were executed on an NVIDIA H100 GPU, averaging 12 hours per model-dataset pair.


\noindent \textbf{Baselines:} We compared MMAS against: (1) \textit{Clean} inputs (no perturbation); (2) \textit{Task-Agnostic UAP (TA-UAP)}~\cite{weng2024learning}, a black-box universal patch without texture scaling; (3) \textit{Texture-Constrained UAP (TC-UAP)}~\cite{huang2024texture}, a vision-only texture-based method adapted for LVLMs.

\noindent \textbf{Evaluation metric:} Success was measured by semantic similarity (cosine distance in embedding space) between the LVLM’s output $\mathbf{r}_{i,j}$ and $\mathbf{r}^*$, averaged across tasks and images: $\text{Similarity} = 1 - d(\mathcal{E}_\phi(\mathbf{r}_{i,j}), \mathcal{E}_\phi(\mathbf{r}^*))$. Higher values indicate stronger attack efficacy. The best performance values for each task are highlighted in \textbf{bold}.

\subsection{Main Results}
\noindent \textbf{Attack performance on different LVLM models and tasks.}
To comprehensively evaluate the performance of our proposed method, we conduct experiments on four LVLM models and three challenging datasets. Table~\ref{tab:Main_table} illustrates the performance of our method, where we utilize the semantic similarity between the target output and the LVLM's output as the evaluation metric. Especially, we choose the target output ``I am sorry'' to avoid the inclusion of high-frequency responses.  We use $s_k=4$. The ``Overall'' column indicates the average semantic similarity score across all tasks.
{Based on this table, we can observe that our attack method consistently achieves the best performance on all models and datasets for different tasks, which shows the effectiveness of our attack method}.


\noindent \textbf{Visualization results.}
As shown in   Figure~\ref{fig:visualization},
we visualize the targeted universal attack. Obviously, each adversarial patch can achieve a universal targeted attack, showing the effectiveness of our learnable perturbations.

\noindent \textbf{Comparison with state-of-the-arts.}
Considering that different LVLM attack methods refer to various settings, we compare our method with each compared method under the same setting for fair comparison. Tables~\ref{tab:compare_to_NIPS} and \ref{tab:compare_to_CroPA} show the comparison results. Obviously, our proposed method significantly outperforms MF-Attack \cite{zhao2024evaluating}  and CroPA \cite{luo2024image}. This is because our  proposed method can effectively update the  visual and textual perturbations by estimating gradients in the victim  model. Moreover, Different from CroPA that only attacks white-box cross-prompt in a single task, our proposed method can attack black-box cross-task inputs for better attack performance. 
As shown in Table \ref{shangye},
we evaluate the performance of our proposed method on realistic LVLM applications Gemini-2.0, GPT-4o and Claude-3.5-Sonnet, where we also achieve best performance.
Also, Table \ref{tab:llavacase} showcases instances of attacks targeting LLaVA. In contrast to state-of-the-art attack methods, our strategy produces greater differences between the model’s responses and the reference captions.



\noindent \textbf{Investigation on the transferability across different datasets and LVLMs.}
Since our proposed attack can craft universal adversarial perturbations applicable to any input across various tasks, examining how well these perturbations transfer is crucial. We present the transfer-attack results in Table~\ref{tab:transferability}, evaluating their effectiveness across diverse datasets and LVLM models. To assess dataset transferability, we create a universal patch targeting the LLaVA model on one dataset, then apply it to the test sets of two other datasets, feeding the results into LLaVA for analysis. For model transferability, we produce a patch against a specific model using the DALLE-3 dataset and test its impact on three additional LVLM models. Our findings reveal that this attack delivers strong performance, underscoring the success of our universal design. However, transferability across datasets proves less effective than across LVLM models, likely due to varying image distributions among the datasets.

\noindent \textbf{Robustness to defenses.} To assess how well our attack withstands protective measures, we test it against three pre-processing defense techniques—Randomization, JPEG Compression, and Diffusion Restoration in Figure \ref{fig:defense}. Unlike compared attack methods, our approach demonstrates greater robustness against these defense methods, as we deliberately design the adversarial perturbation to maximize its disruptive impact. This strategy enhances its ability to steer the LVLM’s reasoning astray, increasing the likelihood of incorrect outcomes compared to a harmless pattern.

\noindent \textbf{Complexity analysis.} To investigate the scalability and practicality of our proposed transfer-attack method, we provide the complexity analysis in Table \ref{complex} on LLaVA-1.5. It indicates that our attack costs relatively fewer GPU resources as our informative constraints are easily achieved with solely loss designs while our samples can achieve better transfer attack performance within single generation process.

\subsection{Ablation Study}

\textbf{Ablation on image and text attacks. } To elucidate the role of each attack of our method, we conduct ablation studies regarding the components (\textit{i.e.}, text attack, image attack and joint optimization).  In particular, we remove each key individual module to investigate its contribution.
1) removing the text attack, 2) removing the image attack and 3) removing the joint optimization. As shown in Table \ref{tab:ablation_attack}, all three module provide the significant performance improvement.
The results demonstrate that each component of our method contributes positively to improving attack performance for different tasks.

\noindent \textbf{Ablation on different target texts.} To show that the success of our proposed attack is not limited to the specific target text ``I am sorry'', we broaden our assessment to include a range of alternative target texts. The experiment features texts of varying lengths and usage frequencies, as detailed in Table~\ref{tab:different_target}. Our findings indicate that the attack excels both overall and in each specific task across these diverse targets, though the similarity of outputs varies depending on the text chosen. 

\noindent\textbf{Ablation study on scaling factor $s_k$}: Table~\ref{tab:ablation_scale} shows similarity peaking at $s_k = 4$, with declines at $s_k = 8$ due to overly fine textures losing coherence, and $s_k = 1$ lacking texture augmentation. Thus, we set $s_k = 4$ in this paper.





\section{Conclusion}
\label{sec:conclusion}

In this paper, we introduced a pioneering framework, MMAS, to expose and exploit the vulnerabilities of Large Vision-Language Models (LVLMs) through coordinated multi-modal adversarial attacks. 
Extensive experiments on multiple datasets show the effectiveness of the proposed attack method for various multi-modal tasks.

{
    \small
    \bibliography{aaai2026}
}

\end{document}